\documentclass[11pt, letterpaper]{berkeley}
\usepackage[all]{hypcap}
\usepackage[authoryear, round]{natbib}
\usepackage{hyperref}[citecolor=lightblue]

\hypersetup{
    colorlinks = true,
}

\usepackage{microtype}
\usepackage{graphicx}
\usepackage{subfigure}
\usepackage{booktabs} % for professional tables
\usepackage{float}
\usepackage{amsmath}
\usepackage{amssymb}
\usepackage{mathtools}
\usepackage{amsthm}
\usepackage{mathrsfs}
\usepackage{nicefrac}
\usepackage{dsfont}
\usepackage{enumitem}
\usepackage{float}
\usepackage{xspace}
\usepackage[capitalize,noabbrev]{cleveref}
\usepackage{subcaption}
\usepackage{wrapfig}
\usepackage{lipsum}
\usepackage{listings}
\usepackage{amsmath}
\usepackage{amssymb}
\usepackage{mathtools}
\usepackage{amsthm}
\usepackage{bbm}
\usepackage{algpseudocode}
\usepackage{setspace}
\usepackage{color}

\definecolor{deepblue}{rgb}{0,0,0.5}
\definecolor{deepred}{rgb}{0.6,0,0}
\definecolor{deepgreen}{rgb}{0,0.5,0}

\newcommand\pythonstyle{\lstset{
basicstyle=\ttfamily\footnotesize,
language=Python,
morekeywords={self, clip, exp, mse_loss, uniform_sample, concatenate, logsumexp},              % Add keywords here
keywordstyle=\color{deepblue},
emph={MyClass,__init__},          % Custom highlighting
emphstyle=\color{deepred},    % Custom highlighting style
stringstyle=\color{deepgreen},
frame=single,                         % Any extra options here
showstringspaces=false
}}

\lstnewenvironment{python}[1][]
{
\pythonstyle
\lstset{#1}
}
{}

\newcommand\pythoninline[1]{{\pythonstyle\lstinline!#1!}}

\makeatletter
\def\mathcolor#1#{\@mathcolor{#1}}
\def\@mathcolor#1#2#3{%
  \protect\leavevmode
  \begingroup
    \color#1{#2}#3%
  \endgroup
}
\makeatother

\usepackage[textsize=tiny]{todonotes}

\Crefformat{equation}{#2Eq.\;(#1)#3}
\Crefformat{figure}{#2Figure #1#3}
\Crefformat{assumption}{#2Assumption #1#3}
\Crefname{assumption}{Assumption}{Assumptions}

\usepackage{crossreftools}
\usepackage{dsfont}
\usepackage{nicefrac}
\usepackage{inconsolata}
\usepackage{algorithm}
\usepackage{amssymb}
\usepackage[skins,theorems]{tcolorbox}

\newcommand{\philipp}[1]{}

\title{{REAL: Benchmarking Autonomous Agents on Deterministic Simulations of Real Websites}}
\reportnumber{} 
\author[1]{Divyansh Garg\textsuperscript{*†}}
\author[4]{Shaun VanWeelden\textsuperscript{*}}
\author[1]{Diego Caples}
\author[5]{Andis Draguns}
\author[2]{Nikil Ravi}
\author[6]{Pranav Putta}
\author[1]{Naman Garg}
\author[7]{Tomas Abraham}
\author[7]{Michael Lara}
\author[7]{Federico Lopez}
\author[1]{James Liu}
\author[1]{Atharva Gundawar}
\author[1]{Prannay Hebbar}
\author[1]{Youngchul Joo}
\author[3]{Jindong Gu}
\author[3]{Charles London}
\author[3]{Christian Schroeder de Witt}
\author[3]{Sumeet Motwani\textsuperscript{†}}

\affil[1]{The AGI Company}
\affil[2]{Stanford University}
\affil[3]{University of Oxford}
\affil[4]{Mercor}
\affil[5]{Contramont Research}
\affil[6]{Plato}
\affil[7]{Independent}

\correspondingauthor{
\mbox{\textsuperscript{*}Equal contribution.} 
\mbox{\textsuperscript{†}Corresponding authors: div@theagi.company, sumeet.motwani@eng.ox.ac.uk.} 
\mbox{Support: real@theagi.company}
}
\begin{abstract}
We introduce \textbf{REAL}, a benchmark and framework for multi-turn agent evaluations on deterministic simulations of real-world websites. REAL comprises high‑fidelity, publicly hosted, deterministic replicas of 11 widely-used websites across domains such as e-commerce, travel, communication, and professional networking. We also release a benchmark consisting of 112 practical tasks that mirror everyday complex user interactions requiring both accurate information retrieval and state-changing actions. All interactions occur within this fully controlled setting, eliminating safety risks and enabling robust, reproducible evaluation of agent capability and reliability. REAL environments are highly configurable, offer complete action/observation space control, and allow researchers to inspect state-changes at any step to define reward signals for training. Our novel evaluation framework combines programmatic checks of website state for action-based tasks with rubric-guided LLM-based judgments for information retrieval. The framework supports both open-source and proprietary agent systems through a flexible evaluation harness that allows research labs to test agentic systems without modification. Our empirical results show that frontier language models achieve at most a $41\%$ success rate on REAL, highlighting critical gaps in current autonomous capabilities. REAL enables easy integration of new tasks, reproducible evaluation, and scalable data generation for post-training web agents. The websites, framework, and leaderboard are available at \url{https://realevals.xyz} and \url{https://github.com/agi-inc/agisdk}.

\end{abstract}

\begin{document}
\maketitle

\section{Introduction}
\label{sec:introduction}
Large Language Models have demonstrated remarkable advances in reasoning capabilities, suggesting a promising path toward human-level performance across domains \citep{kaplan2020scalinglawsneurallanguage, bommasani2022opportunitiesrisksfoundationmodels}. Agents leveraging these models promise to automate countless routine digital tasks with substantial economic impact \citep{brynjolfsson2025generative}, yet consistently struggle with reliably executing multi-turn web interactions that most humans complete effortlessly \citep{xu2412theagentcompany}. Real-world deployment has been slow despite general capability improvements, and can be attributed to the lack of adequate real-world web based training and evaluation environments. This gap not only impedes research progress, but also delays the usefulness of reliably functioning web-agents.
\begin{figure}[ht]
    \centering
    \includegraphics[width=\textwidth]{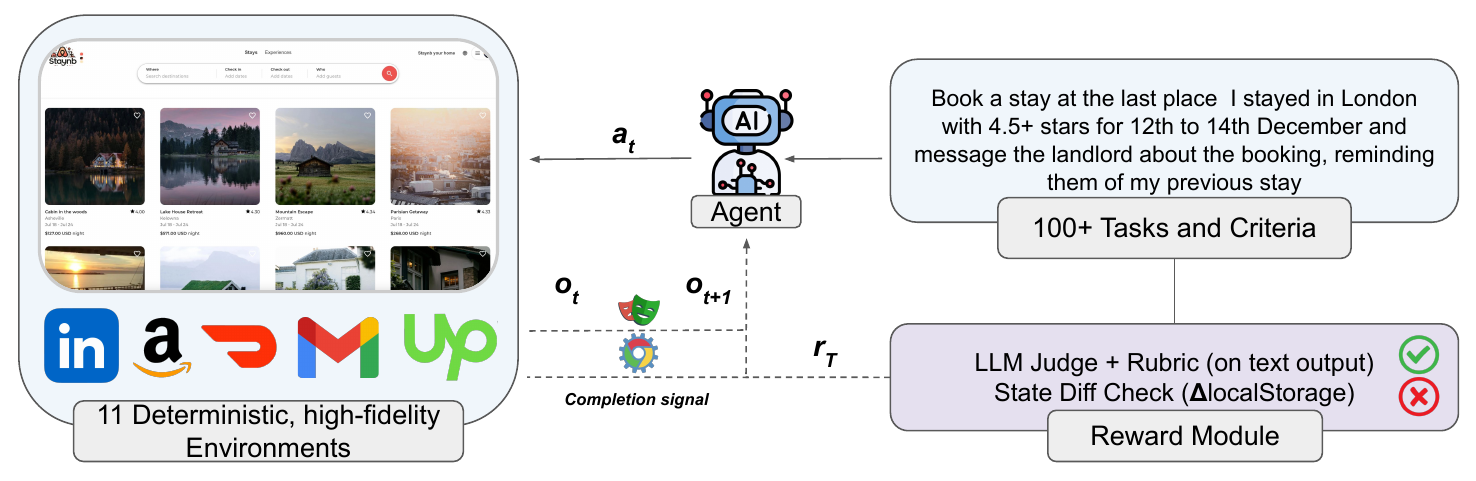}
    \caption{\textbf{The REAL benchmark and framework.} REAL provides 11 realistic, deterministic, high-fidelity web environments (across e-commerce, networking, communication, scheduling, booking, project management) and 110+ evaluation tasks. An agent interacting with the environments receives an observation ($o_t$) and executes actions ($a_t$) to complete a task. Upon completion, an outcome reward ($r_T$) is evaluated via programmatic state verification and/or a rubric based LLM-judge.}
    \label{fig:model-accuracy}
\end{figure}

Current benchmarks for evaluating web agents face several fundamental limitations. First, real websites lack determinism, with constantly changing underlying data and content along with evolving UX workflows, making reproducible evaluation nearly impossible. Second, production websites cannot be configured to test critical edge cases that agents must handle, such as out-of-stock items, network latency variations, or error recovery scenarios \citep{dechezelles2025browsergymecosystemwebagent}. Third, agents may change the state of the website themselves (via payments and state-changes), raising concerns of safety, costs, and robustness during evaluation. Prior works \citep{yao_webshop_2022, zhou2024webarena} have made valuable progress but introduce artificial constraints such as constrained action/observation spaces or simplified tasks and interfaces that may not reflect real-world complexity \citep{yehudai2025surveyevaluationllmbasedagents}. Moreover, these benchmarks are challenging to use as training environments due to the difficulty of defining clear reward signals or observing state-diffs after actions. Lastly, modern websites built with React feature increasingly dynamic and complex structures, making it impractical for web agents to rely on simple HTML extraction methods part of prior benchmarks. These limitations have created a systemic gap between benchmarks and the true challenges of autonomous reliable web navigation.

To address these limitations, we present \textbf{REAL}, a benchmark and evaluation framework designed to test web agents on high-fidelity, deterministic replicas of popular websites. Our approach makes several key advances. First, inspired by WebArena \citep{zhou2024webarena}, we develop accurate representations of 11 widely-used websites (across e-commerce, travel, social media, scheduling) using modern web-development standards. These websites span several pages and mimic the visual and functional fidelity of important real-world websites. We host the sites, reducing the time, cost, and difficulty of self-hosting benchmarks. Second, we make these websites fully deterministic by fixing all data, timestamps, and UX elements while maintaining configurability through URL parameters. This approach allows testing various edge cases (latency, errors, accessibility features) in a reproducible manner while storing website state in the browser's local storage for persistence across sessions.

We provide a flexible test harness that accommodates both open-source and proprietary agentic systems without requiring adherence to a fixed action or observation space. REAL makes it easy to use custom agents by providing unrestricted access to the browser state. This design reflects the current research landscape, where approaches ranging from open APIs \citep{song2025browsingapibasedwebagents} to proprietary \citep{openai2025operator} black-box systems work with custom observation and action spaces. In line with this, we do not impose explicit restrictions on the observation space, allowing users to communicate with the browser via Playwright\footnote{\url{https://playwright.dev/}} for simplicity or Chrome DevTools Protocol (CDP)\footnote{\url{https://chromedevtools.github.io/devtools-protocol/}} for complete control over the browser session.

For evaluations, we provide practical tasks across these websites, covering both information retrieval and state-changing actions \citep{dechezelles2025browsergymecosystemwebagent}. The task input comprises a user request in natural language along with the website configuration URL to initialize the task run with. The REAL framework allocates a persistent CDP session to the agent, enabling low-level browser automation while maintaining state throughout the interaction. When an agent marks the task as complete, it triggers the capture of the local storage state changes and model response. For closed systems, our harness only requires the system to navigate to the start URL before letting the agent execute the task and signal completion by navigating to the final URL with any textual response encoded in the query string. Task performance is evaluated based on two methods: (1) deterministic programmatic comparison of pre-task and post-task browser local storage states for action-oriented tasks \citep{zhou2024webarena}, detecting specific key-value modifications against expected state changes; and (2) a structured LLM-judge for information retrieval tasks \citep{zhuge2024agentasajudgeevaluateagentsagents}, where model responses are evaluated against task-specific rubrics with predefined criteria for correctness and completeness.

We evaluate frontier models with a default agent that we provide as part of REAL. Our current evaluations indicate that no model achieves more than $41.07\%$ performance on our tasks, with Claude 3.7-Sonnet Thinking, Gemini 2.5 Pro Experimental, and OpenAI-o3, and GPT-4o achieving $41.07\%$, $38.39\%$, $34.82\%$, and $14.29\%$ respectively. In this paper, we provide a detailed description of agentic systems and existing benchmarks (Section \ref{sec:related_work}), websites developed for REAL (Section \ref{websites}), how agents can use these environments (Sections \ref{sec:environment} and \ref{sec:method}), our task design and evaluation methodology (Section \ref{sec:experiments}), baseline experimental results (Section \ref{sec:leaderboard}), and implications for future research (Sec. \ref{sec:discussion}).

To summarize, our key contributions include: (1) a collection of 11 deterministic, configurable, high-fidelity simulated web-environments; (2) a flexible evaluation framework supporting both open and proprietary agent systems; (3) a comprehensive set of 112 real-world challenges; (4) a robust evaluation method for each task along with reward signals that could be used for training or synthetic trajectory generation; and (5) an open leaderboard with hosted environments, making agentic evaluations accessible to academia and industry. REAL represents a significant step toward the development and evaluation of highly-capable and reliable real-world web agents.
\section{Motivation and Related Work}
\label{sec:related_work}
\subsection{Benchmarks for Web Agents}
\label{subsec:benchmarks}
% \textcolor{red}{this subsection can be a bit more concise}
Recent advances in large language models (LLMs) have led to growing interest in web agent benchmarks that evaluate an agent’s ability to interact with browser-based environments. Early benchmarks such as MiniWoB \citep{shi2017world} and MiniWoB++ \citep{liu2018reinforcement} established foundational workflows and metrics for evaluating web agents in controlled, reproducible environments. WebShop~\citep{yao_webshop_2022} evaluated agents on their ability to navigate complex e-commerce flows by simulating a single online store. Mind2Web \citep{deng2023mind2web} built on this work, releasing a dataset of more than 2000 open-ended tasks. These benchmarks offer the ability to evaluate agents on pre-defined, offline datasets. Various works have also proposed suits of simulated web environments, for e.g. WebArena \citep{zhou2024webarena} and VisualWebArena \citep{koh2024visualwebarena}. WebArena struggles with realism and task utility, where certain tasks involve artificially constrained ambigious goals or actions that do not reflect everyday web usage \citep{kapoor2024aiagentsmatter}. Moreover, the benchmark requires dedicated hosting infrastructure and overhead, and the environments can be "gamed" \citep{sodhi2024stepstackedllmpolicies} by exploiting shortcuts unavailable in real scenarios.

In addition to benchmarks focusing on everyday tasks, there has also been work focusing on specific use-cases and different dimensions of evaluation. WorkArena~\citep{workarena2024} and WorkArena++~\citep{boisvert2024workarenacompositionalplanningreasoningbased} introduced benchmarks for web agents in the enterprise software setting. AgentBench~\citep{liu_agentbench_2023} is broader in that it includes multiple interactive agentic environments (web browsing, code, gaming, etc.), with the goal of providing insights into more general agent capabilities of LLMs. ST-WebAgentBench~\citep{levy2024stwebagentbenchbenchmarkevaluatingsafety} focused on safety and trustworthiness of web agents, and on assessing web agents’ compliance with organizational policies and safety requirements in enterprise settings.

BrowserGym~\citep{dechezelles2025browsergymecosystemwebagent} offers a unified interface for evaluating agents across multiple existing benchmarks through a standardized observation and action space. BrowserGym's interface forms the foundation for our REAL implementation, which extends its capabilities to address gaps in prior benchmarks (simplified HTML website structures, lack of configurability of environments, tasks that do not fully reflect real-world use-cases, and reproducibility \citep{kapoor2024aiagentsmatter}). 

% However, it also inherits the same fundamental issues outlined for these benchmarks (MiniWoB, WebArena, WorkArena, etc.), such as simplistic tasks or websites that don't represent real-world use cases, a constrained action space, no configurability of the environments, simplified HTML website structures that don't reflect how current websites are developed. \textcolor{red}{Acknowledge BrowserGym as the basis for our RealGym implementation.}

Beyond these specialized benchmarks, efforts to create real-time or live evaluation settings face reproducibility challenges \citep{mcintosh2024inadequacies}. Live websites may change over time, break existing agent behaviors, or introduce unpredictable failures. Conversely, purely synthetic environments—while reproducible—often fail to mirror the complexity and utility of real websites, leading to overfitting and benchmarks that are “solved” but do not generalize \citep{li2024websuitesystematicallyevaluatingweb}. Consequently, there is an unmet need for deterministic, high-fidelity, and readily accessible web benchmarks that support multiple configurations, capture genuine real-world tasks, and act as a testbed for RL research on agentic systems, inspired by foundational work in \cite{openaigym}.

\subsection{Web Agents and Post-training}

A large portion of modern work and everyday tasks is conducted via web-based tools: filling forms, booking food or transport, updating dashboards, retrieving records, ordering items from online shopping sites, or navigating internal portals. Automating even a fraction of these workflows would result in massive economic productivity \citep{brynjolfsson2025generative, bommasani2022opportunitiesrisksfoundationmodels}. Web agents can be customized and made available 24/7; they can also adapt to different kinds of content, making it possible to automate the long tail of web-based tasks that are too cumbersome for humans but too complex for purely software-based automation.

Thus, a new wave of web agents built on top of foundation models has emerged with the development of benchmarks and frameworks described in Section \ref{subsec:benchmarks}. LLM reasoning and planning capabilities in these domains \citep{yao2023react, zhang2024agentohanadesignunifieddata} have led to the deployment of a number of promising agents. AgentQ~\citep{putta2024agentqadvancedreasoning} leverages guided Monte Carlo Tree Search combined with self-critique and iterative post-training to boost multi-step reasoning in complex web navigation tasks. OpenAI’s Operator\footnote{\url{https://cdn.openai.com/operator_system_card.pdf}} and Anthropic's Computer-Use\footnote{\url{https://www.anthropic.com/news/3-5-models-and-computer-use}} employ the companies' respective models to be able to execute simple browser tasks such as ticket booking and form-filling by simulating mouse and keyboard inputs. AgentOccam~\citep{yang2024agentoccamsimplestrongbaseline} improves web task performance by aligning its action and observation spaces with pre-training data.

Several other works along these lines attempt to use exploration and planning to boost performance; WebPilot~\citep{zhang2024webpilotversatileautonomousmultiagent} enhances dynamic web interactions via exploration through a dual-optimized MCTS strategy, and Tree Search for Language Model Agents~\citep{koh2024tree} applies an inference-time best-first search for effective multi-step planning with a value function. Complementing these, WebDreamer~\citep{gu2025llmsecretlyworldmodel} simulates action outcomes to enable speculative planning, collectively demonstrating the improving web-based capabilities of LLM driven agentic systems.

Nevertheless, current agents are still restricted to narrow tasks and have limited error recovery mechanisms, relying on brittle prompts and struggling with complex workflows \citep{yehudai2025surveyevaluationllmbasedagents}. A key reason is the lack of robust training and evaluation environments, which we address in our current work. Ensuring safe, accurate, and efficient navigation is critical in web environments \citep{anwar2024foundationalchallengesassuringalignment, xu2412theagentcompany}, and existing benchmarks sidestep complex UI elements, meaningful environments, realistic tasks, or user input, highlighting the need for a benchmark that can test agents under realistic conditions while ensuring reproducibility.

An emerging paradigm in training foundation models is to post-train them  via reinforcement learning to perform reasoning. Recent releases leveraging test-time compute and reasoning abilities include models such as OpenAI's o1~\citep{openai2024openaio1card} and DeepSeek-R1~\citep{deepseekai2025deepseekr1incentivizingreasoningcapability}. Given the promise of scaling reinforcement-learning in agentic settings, it is increasingly important to have suitable environments to train RL-based agents. REAL generates trajectories and well-defined reward signals for training agents via reinforcement learning. 

\section{REAL Websites}
\label{websites}
% cite: https://www.sciencedirect.com/science/article/pii/S2090447923003003,  Why agents are the next frontier of generative AI, 
REAL consists of 11 high-fidelity realistic website implementations that accurately replicate the functionality and user interfaces of widely-used consumer platforms. We highlight the selection and development process along with several important advantages of our website environments below.
\begin{figure}[htbp]
    \centering
    \includegraphics[width=\textwidth]{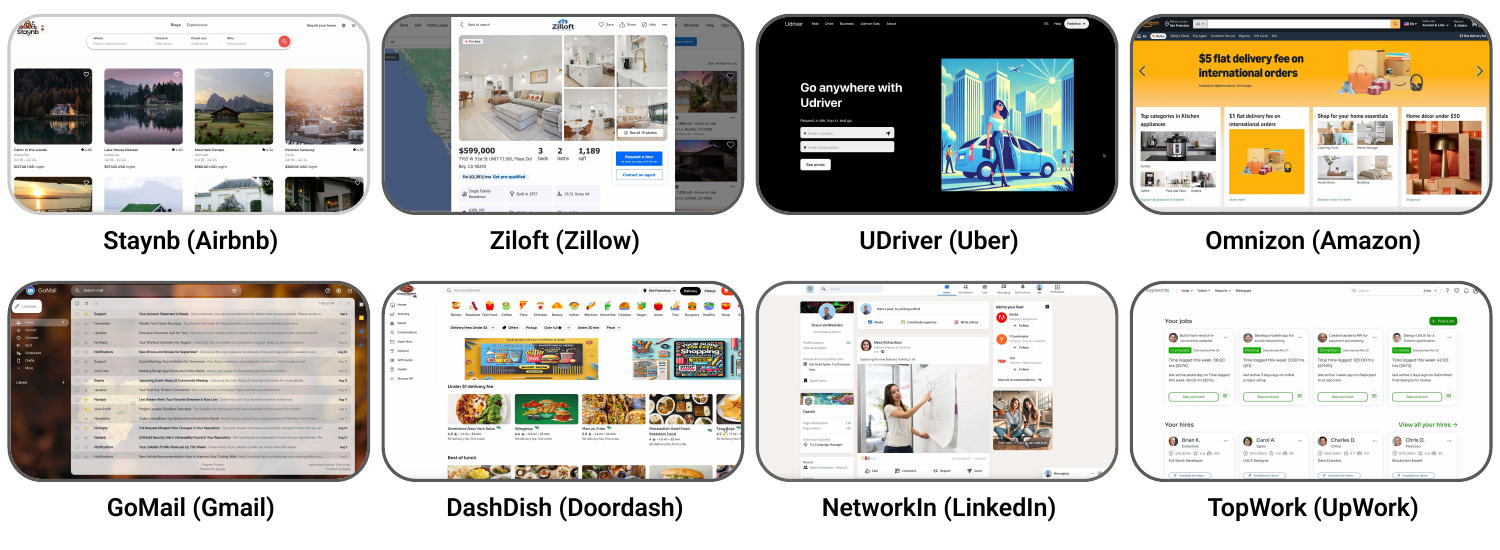}
    \caption{Screenshots of representative web environments included in REAL (8 of 11 shown). These are high-fidelity, deterministic replicas of popular websites, hosted by us for easy accessibility. These environments feature complex, multi-page workflows with persistent state management on the browser, allowing detailed tracking and inspection of state changes induced by agent actions.}
    \label{fig:model-accuracy}
\end{figure}

\subsection{Website Selection}
Our website selection process focused on a diverse set of consumer-facing applications that drive significant web traffic and economic activity \citep{yee2024agents, chui2023economic, handa2025economictasksperformedai}. We identified websites requiring varied interaction capabilities: form completion, reliable online payments, multi-step workflows, dropdown menus, map interfaces, data filtering, information retrieval, and state-dependent elements. The final collection, presented in Table \ref{tab:website-list}, spans key domains including e-commerce, travel, communication, scheduling, freelance marketplaces, property search, etc. This ensures that agents must reliably and accurately handle a representative range of web interactions encountered in everyday tasks—from selecting seats on airline maps to scheduling events and managing payment information—providing comprehensive coverage that allows a systematic evaluation of web agents on important real-world tasks.

\begin{table}[!ht]
    \centering
    \renewcommand{\arraystretch}{1.3}% Increased slightly for readability
    \setlength{\tabcolsep}{6pt}% Adjusted slightly
    \footnotesize
    \begin{tabular}{@{}l l >{\texttt}l p{6.8cm}@{}} % Adjusted width slightly
        \toprule[1pt]
        \textbf{\textsc{Name}} & \textbf{\textsc{Inspired By}} & \textbf{\textsc{REAL URL}} & \textbf{\textsc{Core Functionality}} \\ % Changed header
        \midrule
        Staynb & Airbnb & \href{https://evals-staynb.vercel.app/}{evals-staynb}
            & Search, filter, book, and review vacation rentals; manage bookings. \\ % More action-oriented
        Omnizon & Amazon & \href{https://evals-omnizon.vercel.app/}{evals-omnizon}
            & Browse/search products, manage shopping cart, complete online purchase checkout. \\ % Specific checkout action
        DashDish & Doordash & \href{https://evals-dashdish.vercel.app/}{evals-dashdish}
            & Browse restaurants, customize menu selections, place and manage food delivery orders. \\ % Added manage orders
        GoCalendar & GCal & \href{https://evals-gocalendar.vercel.app/}{evals-gocalendar}
            & Manage calendar views, schedule events, create and modify appointments. \\ % More specific actions
        GoMail & Gmail & \href{https://evals-gomail.vercel.app/}{evals-gomail}
            & Manage inbox (read, label, delete), compose/send emails, handle attachments. \\ % Specific inbox actions
        OpenDining & OpenTable & \href{https://evals-opendining.vercel.app/}{evals-opendining}
            & Search restaurant availability by criteria (time, party size), make/manage table reservations. \\ % Specific criteria
        NetworkIn & LinkedIn & \href{https://evals-networkin.vercel.app/}{evals-networkin}
            & Manage user profile, search for professional connections, view profiles and posts. \\ % Clarified connection features
        UDriver & Uber & \href{https://evals-udriver.vercel.app/}{evals-udriver}
            & Plan trips (set locations), request rides based on service type, view route and fare estimates. \\ % Added service type
        FlyUnified & United & \href{https://evals-fly-unified.vercel.app/}{evals-fly-unified}
            & Search for flights (origin, destination, dates), select seats, book tickets, manage itineraries. \\ % Clarified search params
        TopWork & UpWork & \href{https://evals-topwork.vercel.app/}{evals-topwork}
            & Post jobs (client), search/apply for projects (freelancer), manage proposals and active contracts. \\ % Specified roles
        Zilloft & Zillow & \href{https://evals-zilloft.vercel.app/}{evals-zilloft}
            & Search/filter property listings, save favorites, contact managers, view property details and photos. \\ % Removed vague contact features
        \bottomrule[1pt]
    \end{tabular}
    \caption{\textbf{REAL Website Replicas}: High-fidelity, deterministic clones of popular websites built with modern web frameworks (React, Next.js) for reproducible evaluation of autonomous web agents.}
    \label{tab:website-list}
\end{table}

\subsection{Website Tech Stack}
REAL website environments are implemented using a modern front-end stack centered on React and Next.js. To ensure consistency across environments, each Next.js project utilizes TypeScript and uses the "app" router configuration. User interface components are derived from the Material UI React library. Critically, all websites are publicly deployed via Vercel, ensuring unrestricted internet accessibility without authentication. This public hosting approach eliminates the setup complexities often associated with prior benchmarks requiring local deployment (e.g., via Docker) \citep{koh2024visualwebarena, zhou2024webarena, xu2412theagentcompany}, thereby lowering the barrier for adoption and facilitating wider research community access. This accessibility setup we provide also reflects the likely operational environment for commercial AI systems designed to interact with public web resources \citep{openai2025operator, hu2024dawnguiagentpreliminary, marreed2025enterprisereadycomputerusinggeneralist, wu2023autogen}.

\subsection{Determinism}
To ensure reproducible evaluations, in line with \citep{zhou2024webarena, dechezelles2025browsergymecosystemwebagent}, the websites were designed to be fully deterministic through several key features:
\begin{enumerate}
    \item \textbf{Static Data}: All potentially variable information, such as product prices, availability statuses, and displayed messages, is fixed. This eliminates variability between task executions. AI-generated synthetic data was utilized where appropriate to maintain realism.
    \item \textbf{Predefined Temporal Settings}: Time-dependent elements, including date selectors and time zones, are locked to guarantee consistency across all task runs.
    \item \textbf{Replayability}: As a result, identical task conditions can be reliably recreated, facilitating systematic performance comparisons across different AI systems and experimental configurations.
\end{enumerate}

\subsection{Website Authentication and Browser Management}
To streamline agent interaction, the websites operate in a pre-authenticated state, bypassing standard login procedures and allowing agents immediate access to task-specific functionalities. Similar to \citep{zhou2024webarena}, common anti-automation mechanisms, such as CAPTCHAs or bot detection systems, have been intentionally removed from these environments. Furthermore, website state persists across interactions using the browser's localStorage. This allows for data continuity through page navigation, refreshes, and multi-tab usage, which is similar to realistic user session behavior and enables agents to properly manage stateful tasks \citep{putta2024agentqadvancedreasoning}. Local sessions can be easily cleared by simply navigating to /clear, making REAL environments very convenient to use.

\section{REAL Framework and Environments}
\label{sec:environment}
We model agent interaction within REAL environments as a Partially Observable Markov Decision Process (POMDP). The underlying environment state $s_t \in S$ encompasses the complete browser state at timestep $t$. State transitions $T: S \times A \rightarrow S$ are deterministic, governed by the browser engine executing the website's code in response to agent actions $a_t \in A$. REAL allows agents to interact in two primary ways, allowing users to define the action space $A$ and the observation function $O(s_t) \rightarrow o_t$ accordingly. First, inspired by \citep{yao_webshop_2022, zhou2024webarena, dechezelles2025browsergymecosystemwebagent}, high-level interactions use Playwright, which involves an action space $A$ typically comprising user-level commands (see Section \ref{actions}) and an observation space $O$ (screenshots, full DOM, or the accessibility tree) which can be configured and is explained in Section \ref{obs}. For lower-level control, we also provide an option for direct access to a Chrome DevTools Protocol (CDP) session, which enables a near-unrestricted action space $A$ consisting of any valid CDP command and allows for a richer observation space $O$ consisting of the entire live browser session state.

At each step $t$, the agent receives observation $o_t \in O$, selects action $a_t$ conditioned on the task $i$ and potentially the history $(o_1^{t}, a_1^{t-1})$, leading deterministically to the next state $s_{t+1}$ and observation $o_{t+1}$. Task success ($r=1$) or failure ($r=0$) is determined by an outcome reward function $r$, evaluated only at the final timestep $T$. In Section \ref{sec:method}, we describe the agent harness and evaluation flow.

\subsection{Observation Space}
\label{obs}
REAL offers configurable observation spaces $O(s_t) \rightarrow o_t$ which can be specified based on an agent's chosen interaction modality (high-level Playwright or low-level CDP).

For agents interacting via the high-level Playwright interface, we provide default agents that can be configured to use an observation space $O$ including one or more of the following components: \textbf{Screenshots}, visual renderings of the current web page; \textbf{Full DOM}, the complete Document Object Model structure of the page; \textbf{Accessibility Tree}, a representation of the page structure based on accessibility APIs, providing semantic information about elements. This is broadly consistent with other web benchmarks and high-level actions existing agents use to operate \citep{openai2025operator, putta2024agentqadvancedreasoning, koh2024tree, yang2024agentoccamsimplestrongbaseline}. Alternatively, REAL provides the agent with direct access to the Playwright \texttt{Browser} object itself, allowing the use of information derivable through the Playwright API as its observation space.\footnote{\url{https://playwright.dev/docs/api/class-browser}}

For agents requiring fine-grained control through the low-level Chrome DevTools Protocol (CDP), the observation space encompasses the entire live browser session state accessible via the CDP connection. This provides maximum flexibility, allowing the agent to observe any aspect available as part of the browser's current session.\footnote{\url{https://chromedevtools.github.io/devtools-protocol/}} This flexibility enables researchers to adapt the observation space to the specific input requirements and capabilities of custom agent architectures or scaffolds.

\subsection{Tasks and Action Space}
\label{actions}
Agents interact with the environment by selecting actions \( a_t \) from an action space \( A \) to accomplish evaluation tasks detailed in Section \ref{sec:experiments}. Our goal is to keep the definition of \( A \) highly flexible \citep{zhang2024webpilotversatileautonomousmultiagent} and dependent on the chosen interaction setup.

When using the Playwright interface, the action space \( A \) consists of high-level commands simulating standard user inputs. This includes but is not limited to operations such as text input, manipulation of checkboxes, mouse clicks, keyboard commands and shortcuts, file uploads, focus elements, drag and drop, and scrolling.\footnote{\url{https://playwright.dev/docs/input}} This allows agents designed around user-level actions to operate naturally within REAL environments, though \( A \) is not strictly limited to these examples.

Agents interfacing with environments via CDP (\cite{dechezelles2025browsergymecosystemwebagent, workarena2024} also incorporate rich observation spaces with CDP) have access to a substantially broader action space. This low-level control permits a wide range of interactions directly within the browser environment. For instance, agents can execute commands for direct Document Object Model (DOM) modification, arbitrary JavaScript execution within the page's context, performance profiling, emulation of different devices or network conditions, interception and modification of network requests, and even detailed browser session debugging using tools like breakpoints.
% High-level actions via playwright, inluding but not limited to as Text input, checkboxes, select optoins, mouse click, keys and shortcuts, upload files, focus element, drag and drop, scrolling. 
% \footnote{\url{https://playwright.dev/docs/input}}. However, the goal of RealGym is flexibility and this applies to the action space too. Therefore, with CDP agents can use low level commands that can even modify the DOM, script execution to run javascript code, performance profiling, emulation, network control, browser session debugging with breakpoints, etc anything else that is relevant.

% List of actions and how they must be executed. Similar to how WebArena lists the action space and intents of tasks. We let the developers do whatever it is that they want to do with the chrome cdp session. We can divide this into low level actions i.e. the cdp commands such as commands to modify the dom, input simulation for clicks, inputs, scrolling, script execution to run javascript code, network control, debugging. High level actions can then be used as browser base commands - user interface interactions and compound commands simulation multiple low-level actions. Our tasks are information retrieval tasks and action-oriented tasks. We provide more details about these tasks in Section \ref{sec:experiments}.

% playwright launches the browser. interactive.py headless or not headless mode for the browser. playwright lets the human interact with the browser. Anything in playwright. 

\subsection{Rewards}
Our framework primarily uses an outcome reward function $r \in \{0, 1\}$ to evaluate task success upon completion (at timestep $T$). This binary outcome reward indicates whether the agent successfully achieved the specified task goal $i$ and is determined as follows (see Section \ref{sec:experiments} for details on Action-based and Information Retrieval tasks, similar to \cite{zhou2024webarena}):
\begin{itemize}
    \item \textbf{Action-based Tasks ($r_A$):} Rewards are determined by programmatic verification function $f_{eval}$, which compares the difference between the initial ($s_0$) and final ($s_T$) `localStorage` states against a set of predefined key-value assertions specific to the task goal $i$. $r_A=1$ if and only if all assertions pass with an exact match.
    \item \textbf{Information Retrieval Tasks ($r_R$):} Rewards are determined by an LLM-judge evaluation function $g_{eval}$, which assesses the agent's final submitted text response against a pre-determined task-specific rubric. $r_R=1$ if the response is judged as correct according to the rubric.
    \item \textbf{Combined Tasks:} Require both $r_A=1$ and $r_R=1$ for the overall task reward $r$ to be 1.
\end{itemize}
We note that while the current version of REAL provides binary outcome rewards, the underlying framework components (deterministic environment, state tracking via `localStorage`, programmatic checks) are flexible enough to support the definition and use of dense, step-wise reward functions for reinforcement learning \citep{lightman2023letsverifystepstep, putta2024agentqadvancedreasoning}.

\subsection{Evaluation Functions}
\label{configendpoint}
REAL offers several endpoints for evaluations, debugging, and environment configurations:
\begin{itemize}
    \item \textbf{\texttt{/config}:} Used to initialize the environment for a specific task run. Appending query parameters to this endpoint allows setting both universal and website-specific configurations (detailed in Section~\ref{sec:configurable_environments}), such as simulated \texttt{latency}, error mode flags (e.g., \texttt{error\_finding\_driver}), accessibility settings (\texttt{hide\_aria\_labels}), and run identifiers (\texttt{run\_id}, \texttt{task\_id}).
    
    \item \textbf{\texttt{/submit}:} The agent must navigate to \texttt{/submit} to signal task completion for leaderboard submissions. This action captures the final \texttt{localStorage} state and the agent's textual response. This captured data is then used by the evaluation harness to compute the reward~$r$ and record the result on the public leaderboard associated with the provided \texttt{run\_id}. 
    
    \item \textbf{\texttt{/finish}:} Whenever the website state changes, those changes are saved in the website \texttt{localStorage} state. Navigating to \texttt{/finish} at any point displays the difference between the initial state and current state, allowing users to inspect the precise state changes.
        
    \item \textbf{\texttt{/clear}:} Navigating to \texttt{/clear} resets the website's \texttt{localStorage} to its default empty state.
\end{itemize}

% \subsection{High-level use and implementation}
% How is everything implemented and hosted. How does the agent broadly interact with pages. Exact RL terminology for agent interaction with the web, how history is stored, how the reward functions work. What should the agent's output look like and how is it used with the benchmark? \textcolor{red}{Figure depicting full usage}
\section{Evaluation Tasks}
\label{sec:experiments}
REAL consists of a suite of 112 evaluation tasks across 11 website environments. These tasks are designed to assess performance on realistic, multi-turn interactions that mirror common user goals and workflows encountered on the internet. These tasks go beyond simple, atomic actions and are assigned difficulty levels (easy, medium, and hard), providing a relative indication of factors such as amount of planning required, number of interaction steps, constraints, or the required reasoning depth. These tasks involve both information seeking and state manipulation within the environments, employed early on by \cite{yao_webshop_2022, zhou2024webarena, koh2024visualwebarena, yoran2407assistantbench, liu2023agentbenchevaluatingllmsagents}. Each task is based on a natural language instruction (the `goal') provided to the agent, potentially accompanied by specific environment configurations set via the `/config' endpoint (as described in Section \ref{sec:configurable_environments}). We categorize tasks as follows.

\subsection{Information Retrieval Tasks}
Information Retrieval tasks require the agent to navigate an environment, locate specific pieces of information, potentially merge findings from multiple locations, and report the result \citep{deng2023mind2web, zhou2024webarena}. The goals range in complexity from simple lookups on a single page (e.g., identifying the first few items listed, finding a specific flight time) to more complex queries requiring navigation across pages or filtering based on constraints (e.g., finding the number of restaurants matching a specific category, summarizing event counts across different calendars for a given month).

Evaluation for only retrieval tasks is based on the final text response generated by the agent upon task completion (submitted via the `/submit' endpoint). An LLM judge \citep{zheng2023judging} evaluates the agent's response against a predefined task-specific `rubric` to determine if the retrieved information is accurate and complete according to the ground truth present in the environments.

\subsection{Action-based Tasks}
Action-based tasks require an agent to perform actions that modify the environment's state. These tasks represent common goal-oriented web usage, such as booking a flight or ride, scheduling calendar events, filling out forms with specific details, professional networking, etc. These tasks often require interpreting complex instructions involving multiple constraints (e.g., specific dates, times, locations, passenger numbers, item types, payment details) \citep{yao_webshop_2022}.

The evaluation of action-based tasks relies on programmatic verification of the final website stage, captured via the browser's `localStorage' when the agent navigates to `/submit'. We use `state-check' mechanisms to inspect the difference between the initial and final `localStorage` state \citep{zhou2024webarena}. A task is considered complete only if all specified state conditions are met. This provides an objective and deterministic measure of the agent's ability to effect precise state changes.

\subsection{Combined Tasks and Additional Details}
Several tasks within REAL combine elements of both retrieval and action (identified as `challengeType: retrieval-action'). For instance, an agent might be asked to find the price of an item, add it to the cart, complete the purchase, and then report the final cost. Furthermore, similar to \cite{zhou2024webarena}, we specifically include tasks designed to be impossible under the given deterministic conditions (`possible: false'), such as attempting to book a flight that doesn't exist or using deliberately invalid payment information. These tasks allow us to evaluate an agent's ability to recognize failure conditions, potentially utilize error recovery strategies (if applicable), and accurately report the inability to complete the requested goal, rather than hallucinating success or failing silently \citep{li2024websuitesystematicallyevaluatingweb, renze2024self, kara2025waber}. Combined, our evaluations test important aspects of agent performance and reliability in real-world scenarios.

\section{Agent Harness}
\label{sec:method}
The REAL Agent Harness provides a standardized interface for evaluating varied agent implementations \citep{yao2023react, openai2025operator, putta2024agentqadvancedreasoning, shinn2023reflexion, koh2024tree, yang2024agentoccamsimplestrongbaseline} with minimal required modification. Our goal is to prioritize simplicity and compatibility, enabling researchers to evaluate agents across multiple interaction paradigms while maintaining their existing agent architectures. This approach reduces the technical overhead associated with benchmarking, promoting broader adoption and research across academia and industry.

\subsection{Technical Architecture}
The harness offers three integration settings to accommodate different types of agent architectures \citep{acharya2025agentic}. As discussed in Section \ref{sec:environment}, direct Playwright integration grants the user access to a Playwright Browser instance, which enables high-level control of BrowserContext and Page objects for standard web interaction primitives (navigation, element interaction, DOM inspection). For agents requiring lower-level control, the harness provides a WebSocket endpoint for the Chrome DevTools Protocol (CDP), which allows direct execution of CDP commands across domains like DOM, Runtime, Network, and Input for fine-grained state manipulation. Third, for agents employing black-box systems, our harness supports integration via URL endpoints that expose the browser instance, allowing external controllers to attach and manage the session.

\subsection{Evaluation Flow}
A task is initialized when the harness receives a task definition, including a natural language goal ($i$) and a configuration URL. The harness launches and manages a dedicated browser instance, navigating it to the specified /config endpoint (Section \ref{configendpoint}). Subsequently, control is passed to the agent via its selected integration setting and the agent then enters an iterative loop, receiving observations $o_t$ and executing actions $a_t$ which get translated to corresponding API calls (e.g., page.click(), page.evaluate(), or Input.dispatchKeyEvent via CDP). This interaction cycle continues until the agent attempts to fulfill the task goal $i$, potentially constrained by a maximum step limit. Task completion is signaled by the agent navigating to the designated /submit endpoint or just returning an output/ending the loop (for local client-side evaluation). The harness intercepts this final step, and captures two primary details: the final localStorage state and any agent generated text response (optional), passed via the URL query string. These are then programmatically passed to the task-specific evaluation (outcome reward) functions we describe in Section \ref{sec:experiments}. 

\subsection{Submissions and Leaderboard API}
\label{submissions}
Our evaluation framework operates in two modes. A local evaluation returns results directly, allowing quick iterative development and debugging. Researchers can also use the /finish endpoint during these runs to inspect the intermediate localStorage state-diff without concluding the evaluation. Alternatively, navigating to /submit with the correct id can be used for a formal leaderboard submission attempt. This initiates a full evaluation for potential inclusion in public rankings, subject to manual verification. This supports both private research and verified public benchmarking, similar to \citet{zhou2024webarena,yoran2407assistantbench, workarena2024, wang2025computer}.

\subsection{Integration of Custom Agents}
The REAL harness is designed as an adapter layer to minimize the effort required to integrate custom agents. Researchers can connect their existing systems, including those with proprietary reasoning or planning modules, by implementing the interaction logic against just one of the provided interfaces (Playwright API, CDP command execution, or external control via URL access). This significantly reduces the need for agent-side architectural modifications, lowering the barrier to participation for academic and commercial teams while enabling standardized benchmarking.

\section{Configurable Environments} \label{sec:configurable_environments}
REAL incorporates a configuration framework that enables precise control over testing conditions, significantly improving its utility for rigorous agent evaluations. Addressing limitations of static environments found in prior benchmarks \citep{yao_webshop_2022, zhou2024webarena, liu_agentbench_2023} and the non-reproducibility of live websites \citep{kapoor2024aiagentsmatter}, REAL implements a two-level configuration system—universal and website-specific. This structure supports systematic evaluations to develop reliable agents \citep{dechezelles2025browsergymecosystemwebagent, workarena2024}, while maintaining the determinism crucial for reproducible results. Configurations are applied for each task run via standard query string parameters appended to a dedicated /config endpoint on each website.

Universal configurations apply globally across all websites within the benchmark, and are used to establish consistent baseline conditions. Parameters at this level include settings such as simulated network \texttt{latency} (default 2000ms), the \texttt{hide\_aria\_labels} flag (default \texttt{false}) to control the presence of ARIA attributes for accessibility testing, and identifiers for experimental management (\texttt{run\_id}, \texttt{task\_id}). Configuring these parameters allows researchers to isolate and understand the effects of broader web/browser based factors on agent behavior across tasks and websites \citep{kapoor2024aiagentsmatter}.

Website-specific configurations allow  granular control over the internal state, behavior, and simulated backend processes tailored to each individual application. This capability is essential for simulating specific operational scenarios, user contexts, and edge cases related to each site. Beyond initializing basic states, for example the \texttt{total\_conversations} on GoMail, these parameters provide detailed control relevant to real-world usage \citep{krishnan2025ai, xu2412theagentcompany}. Here, we use the UDriver environment as an illustration of the website-specific parameters researchers can configure:
\begin{itemize}
    \item \textbf{Introduce controlled error states} into workflows to evaluate agent error detection and recovery capabilities (e.g., setting \texttt{error\_finding\_driver=true} or \texttt{error\_booking\_ride=true}).
    \item \textbf{Modify the timing or latency of operations} to assess agents under different system response times (e.g., adjusting \texttt{simulating\_searching\_driver\_delay} or \texttt{simulating\_booking\_trip} durations).
    \item \textbf{Modify application-specific logic parameters}, such as internal pricing calculations or discount availability (e.g., modifying \texttt{udriverx\_multiplier} or \texttt{comfort\_discount}).
    \item \textbf{Set initial content or regional contexts} via data presets (e.g., using \texttt{location\_preset=2} to initialize the environment with data relevant to New York).
\end{itemize}
This fine-grained control enables targeted evaluations focused on how agents handle specific website behaviors or data conditions. Detailed configurations for each environment part of REAL are provided on our website\footnote{See \url{https://www.realevals.xyz/websites/udriver} with the appropriate site name for specific configuration details.}.

This dual-level configuration system in REAL provides researchers an extensive amount of control over specific experimental variables within a deterministic framework. This allows for a systematic evaluation of agent reliability across several general and task/website specific modifications \citep{xia2025evaluationdrivendevelopmentllmagents}. The ability to precisely define, reproduce, and study such conditions allows for research that is often infeasible on live, dynamic websites or less flexible benchmark environments.
\section{Leaderboard}
\label{sec:leaderboard}
We evaluated our baseline agent with a large set of frontier models on our REAL environments. This section presents the quantitative performance and discusses some important observations derived from analyzing agent interaction trajectories.
\begin{figure}[htbp]
    \centering
    \includegraphics[width=0.97\textwidth]{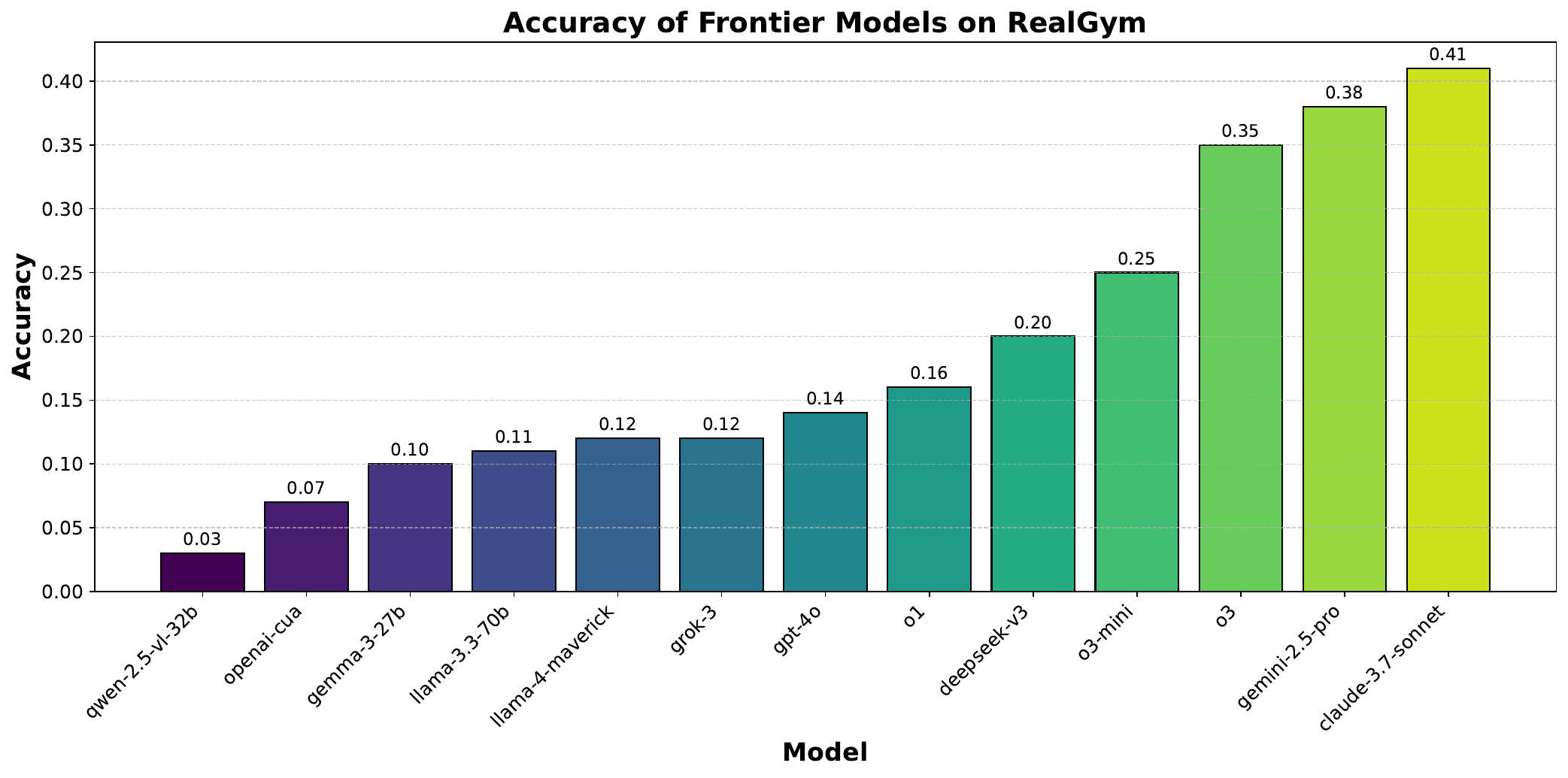}
    \caption{Performance of evaluated models on the REAL benchmark, measured by end-to-end task success rate of our baseline agent across 112 tasks. Claude 3.7 Sonnet-Thinking achieves 41.07\%.}
    \label{fig:model-accuracy}
\end{figure}

\begin{figure}[htbp]
    \centering
    \includegraphics[width=0.92\textwidth]{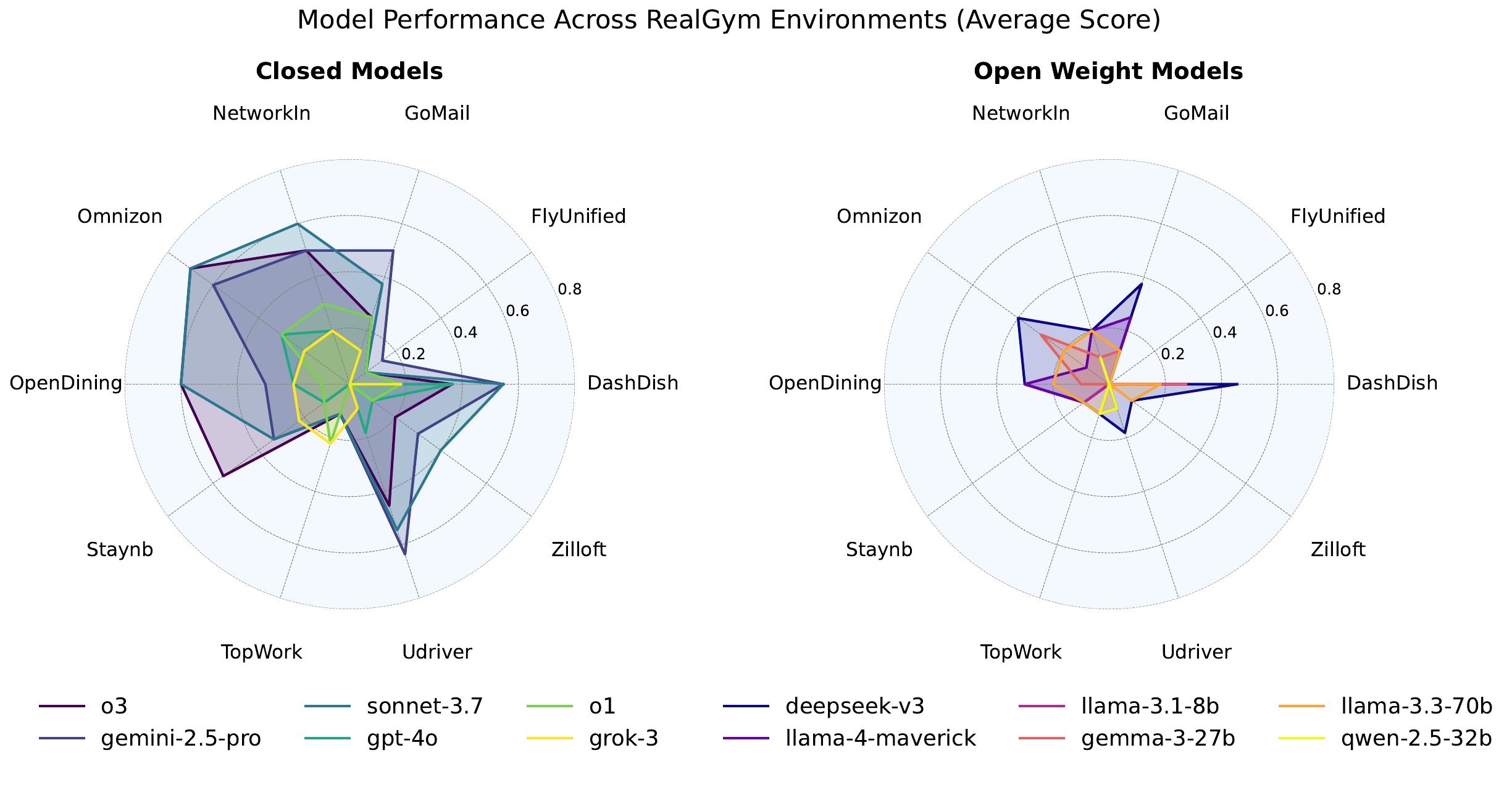}
    \vspace{-1px}
    \caption{A per-website performance breakdown for several frontier models across REAL environments. TopWork and FlyUnified are consistently the most challenging environments.}
    \label{fig:model-accuracy2}
\end{figure}

The overall end-to-end task success rates across the 112 REAL tasks for various models are summarized in Figures \ref{fig:model-accuracy} and \ref{fig:model-accuracy2}. Performance varied considerably across tested models. The current leading model is Claude-3.7-Sonnet-Thinking\footnote{\url{https://assets.anthropic.com/m/785e231869ea8b3b/original/claude-3-7-sonnet-system-card.pdf}}, achieving a success rate of $41.07\%$, followed by Gemini-2.5-Pro-Experimental at $38.39\%$. Similarly, other reasoning models also perform much better than standard pre-trained models, for example o3 ($34.82\%$), o3-mini ($25.00\%$), and o1 ($16.07\%$)\footnote{\url{https://cdn.openai.com/o1-system-card-20240917.pdf}}. Despite this, there is a significant room for improvement. Open source models currently lag behind, with Llama-4-Maverick ($12.50\%$) showing effectively similar performance to Llama 3.3 70B ($10.71\%$) \citep{grattafiori2024llama3herdmodels}. This suggests that increases in scale alone, at least between these specific models, did not translate into improved practical web navigation capabilities. Notably, DeepSeek V3 ($19.64$\%) \citep{deepseekai2025deepseekr1incentivizingreasoningcapability} show much better performance than Llama models. Small models lag significantly, with Llama-3.1-8B, Qwen-2.5-vl-32B, and Gemma-3-27B achieving only $1.79\%$, $2.68\%$, and $9.82\%$ respectively, underscoring the requirement for substantial model capacity and training to handle the complexities of agentic performance. We also evaluated OpenAI's Computer-Using Agent (CUA) model \citep{openai2025operator}, recording a success rate of only $7.14\%$. When these trajectories were manually examined, we observed that CUA was frequently distracted by irrelevant details and did not complete the final steps of several tasks.

Overall, our results demonstrate that reliable, autonomous navigation of websites and completion of tasks remains a significant challenge for current frontier models. Similar results are also observed across benchmarks, as studied by \cite{dechezelles2025browsergymecosystemwebagent}. We do expect performance to go up with better agent scaffolds beyond our baseline, that integrate search and post-training similar to \citet{putta2024agentqadvancedreasoning, koh2024tree, su2025learnbyinteractdatacentricframeworkselfadaptive}. REAL is flexible enough to develop harder tasks on the same environments if agents saturate the current test-set in the short term.
\subsection{Qualitative Observations}
We analyze interaction traces and outline two common failure modes contributing to low performance of models with the baseline agent on our benchmark.

\textbf{Inadequate Failure Recognition and State Verification} Agents often fail to assess whether they have successfully completed all parts of the task, lending more weight to their perceived previous actions than the actual updated observation space. For example, within Omnizon (e-commerce), an agent tasked with adding two items to the cart might add the first item but then fail to add the second item due to clicking an incorrect button or misinterpreting the page state. Despite the cart only containing one item, the agent proceeds through checkout, concluding the interaction under the false assumption that the task was complete. Immediate state-verification against the overall goal and error direction thus remain challenging, as also observed in \citet{zhou2024webarena, li2024websuitesystematicallyevaluatingweb}.

\textbf{Navigation Dead Ends and Lack of Recovery} Agents often struggle when encountering non-standard navigation flows or unexpected states, and lack the intuitions to backtrack effectively. For example, in the Udriver ride-booking environment, an agent might correctly initiate a booking but then click on an option to schedule the ride for a future time, entering a sub-menu. Once in this sub-environment, agents frequently fail to identify the correct UI element (e.g. a back button, cancel option, or the intended next step) to return to the primary task. Interpreting and understanding the purpose of UI elements to apply reliable exploration or backtracking strategies remains an issue, and agents enter loops of clicking irrelevant elements, effectively getting stuck.
\section{Discussion and Future Work}
\label{sec:discussion}
In this work, we introduced REAL, a benchmark and framework designed to evaluate and improve the accuracy and reliability of autonomous web agents. By providing 11 high-fidelity, deterministic web environments along with 112 realistic multi-turn tasks, REAL offers important improvements over prior benchmarks \citep{yao_webshop_2022, zhou2024webarena, dechezelles2025browsergymecosystemwebagent}. Furthermore, the flexible agent harness, supporting both high-level (Playwright) and low-level (CDP) interaction for open and proprietary systems, alongside publicly hosted environments and a leaderboard, lowers the barrier to entry and facilitates standardized, comparative research. Our findings on the benchmark show the challenges posed by realistic environments and highlights the substantial room for improvement in agent capabilities on consequential web tasks \citep{li2024websuitesystematicallyevaluatingweb}.

Beyond its primary role as an evaluation benchmark, REAL is designed to serve as a valuable environment for data generation and agent post-training \citep{shinn2023reflexion, putta2024agentqadvancedreasoning, zhou2025sweetrltrainingmultiturnllm}. The deterministic nature of our environments and detailed state information (especially precise state changes resulting from actions as well as a rich observation space via complete CDP session access), and well defined outcome reward functions allow for the collection of interaction trajectories suitable for various post-training approaches, including imitation learning \citep{zelikman2022starbootstrappingreasoningreasoning, chen2023fireactlanguageagentfinetuning} and reinforcement learning \citep{schulman2017proximalpolicyoptimizationalgorithms, putta2024agentqadvancedreasoning, qi2025webrltrainingllmweb}. Researchers can readily extend the benchmark by defining new tasks with custom goals and evaluation metrics tailored to specific training objectives. The depth and complexity of our websites and tasks makes REAL particularly relevant for advancing RL techniques aimed at improving the reasoning and planning capabilities \citep{xiang20252reasoningllmslearning, jaech2024openai} of web agents. While REAL is currently limited to only outcome rewards and a small suite of evaluation tasks, future iterations will include a dedicated library \citep{openaigym} and set of training tasks to streamline RL post-training workflows \citep{kumar2025llmposttrainingdeepdive, deepseekai2025deepseekr1incentivizingreasoningcapability}. The framework is flexible enough to allow the use of advanced planning \citep{gu2025llmsecretlyworldmodel}, multi-agent \citep{motwani2025maltimprovingreasoningmultiagent}, or tree-search methods \citep{koh2024tree}; future work will also focus on providing improved integration support for such search-based agent architectures.

REAL is designed for extensibility, and the task suite can be expanded with scenarios requiring more sophisticated long horizon reasoning \citep{chen2025reinforcementlearninglonghorizoninteractive} or cross-application workflows \citep{workarena2024, bonatti2024windowsagentarenaevaluating, xu2412theagentcompany} as agents improve. We acknowledge current limitations, including the finite set of 11 environments and the focus specifically on web-based interactions, which represent only a subset of potential agent applications. However, REAL delivers an important, rigorous, and accessible framework designed to bridge the gap between current research and practical deployment. Our goal is to drive the development of autonomous agents \citep{putta2024agentqadvancedreasoning}, and REAL provides the benchmark and framework necessary to evaluate and train these systems to improve their capability and reliability for important real-world applications.

\section*{Acknowledgments}
We would like to thank Milind Maiti, Harshit Sikchi, Julia Kiseleva, Dylan Bowman, Jack Bai, Andrew Gritsevskiy, Matthew Tang, and Lucas Vium for valuable discussions. The websites and configurations are designed under the principles of fair use to serve as tools for research and development.

% \newpage
\bibliography{main}

\appendix
\newpage
\subsection*{Disclaimer}
We will aim to keep improving the benchmark, test suite, and training environment in the near future and have strived to acknowledge the enormous strides made by past work in the area. We are not affiliated, associated, authorized, endorsed by, or in any way officially connected with the real-world companies, brands, or entities represented by the mimicked websites. All company names, logos, and trademarks used on the Platform belong to their respective owners. Results and evaluations conducted on the Platform are for testing and benchmarking purposes only and should not be construed as equivalent to performing actions on actual websites or applications. While we strive to provide realistic simulations, we do not guarantee the accuracy, completeness, or currency of the content or workflows presented in the mimicked websites. The websites and configurations are designed under the principles of fair use to serve as transformative tools for research. Any similarities to real-world counterparts are intended only to replicate core interaction flows in a controlled environment and do not represent the full functionality or appearance of the actual websites.

\end{document}